\documentclass[letterpaper, 10 pt, conference]{ieeeconf}  

\IEEEoverridecommandlockouts                              

\usepackage{amssymb}
\usepackage{graphicx}
\usepackage{hyperref}
\usepackage{balance}
\usepackage{tabularx}
\usepackage{multirow}
\usepackage{multicol}
\usepackage{booktabs}
\usepackage{caption}
\usepackage{mathtools}
\usepackage{siunitx}
\usepackage{amsmath}
\usepackage{MnSymbol}
\usepackage{kotex}
\usepackage{graphicx}
\usepackage{subcaption}
\usepackage{tablefootnote}
\usepackage{bbding}
\usepackage{threeparttable} 
\usepackage{cite}
\hypersetup{
    colorlinks=False, 
    urlcolor=black,
}
\usepackage{makecell}
\usepackage[table]{xcolor}
\definecolor{lightred}{RGB}{255, 200, 200}
\definecolor{lightorange}{RGB}{255, 230, 180}
\definecolor{lightyellow}{RGB}{255, 255, 200}
\definecolor{lightgreen}{RGB}{200, 255, 200}

\title{\LARGE \bf
G\textsuperscript{2}S-ICP SLAM: Geometry-aware Gaussian Splatting ICP SLAM
}

\author{Gyuhyeon Pak$^{1}$, Hae Min Cho$^{2}$ and Euntai Kim$^{1, *}$
\thanks{$^*$ is that the corresponding author.}
\thanks{$^{1}$Gyuhyeon Pak and Euntai Kim are with the Department of Electrical and Electronic
Engineering, Yonsei University, Seoul 03722, Republic of Korea
        {\tt\small \{gh.pak, etkim\}@yonsei.ac.kr}}
\thanks{$^{2}$Hae Min Cho is with the School of Computing, Gachon University, Sujeong-gu, Seongnam-si, Gyeonggi-do 13120, Republic of Korea
        {\tt\small hmcho9@gachon.ac.kr}}
}

\begin{document}

\maketitle
\thispagestyle{empty}
\pagestyle{empty}

\begin{abstract}
In this paper, we present a novel geometry-aware RGB-D Gaussian Splatting SLAM system, named G\textsuperscript{2}S-ICP SLAM. The proposed method performs high-fidelity 3D reconstruction and robust camera pose tracking in real-time by representing each scene element using a Gaussian distribution constrained to the local tangent plane. This effectively models the local surface as a 2D Gaussian disk aligned with the underlying geometry, leading to more consistent depth interpretation across multiple viewpoints compared to conventional 3D ellipsoid-based representations with isotropic uncertainty. To integrate this representation into the SLAM pipeline, we embed the surface-aligned Gaussian disks into a Generalized ICP framework by introducing anisotropic covariance prior without altering the underlying registration formulation. Furthermore we propose a geometry-aware loss that supervises photometric, depth, and normal consistency. Our system achieves real-time operation while preserving both visual and geometric fidelity. Extensive experiments on the Replica and TUM-RGBD datasets demonstrate that G\textsuperscript{2}S-ICP SLAM outperforms prior SLAM systems in terms of localization accuracy, reconstruction completeness, while maintaining the rendering quality.

\end{abstract}

\vspace{6pt}


\section{INTRODUCTION}
Visual Simultaneous Localization and Mapping (vSLAM) has been a fundamental technique of robotics and augmented reality applications, offering efficient and scalable solutions for camera tracking and scene reconstruction. Traditional SLAM pipelines~\cite{orb-slam, orbslam2, orb-slam3, qin2018vins, forster2014svo, cho2021spslam} primarily rely on sparse and geometry-driven representations such as keypoints, landmarks, or surfel maps. While these approaches achieves computational efficiency and robustness, they often fall short in delivering dense, photorealistic reconstructions and struggle with geometric consistency across viewpoints. This limitation restricts their applicability in scenarios requiring perceptually meaningful and structurally accurate maps. 

Recent advancements in radiance field-based representations~\cite{mildenhall2021nerf, gaussiansplatting}, particularly 3D Gaussian Splatting (3DGS), have demonstrated the potential for real-time, high-fidelity rendering of complex scenes. These methods synthesize view-dependent RGB images by modeling scene point as a volumetric Gaussian, which enables impressive photorealism. However, their isotropic and volumetric nature often introduces depth inconsistencies when observed from varying viewpoints. in SLAM applications, this lack of multi-view geometric coherence can degrade pose accuracy and lead to spatial artifacts.

\begin{figure}[t]
\centering
\includegraphics[width=0.98\columnwidth]{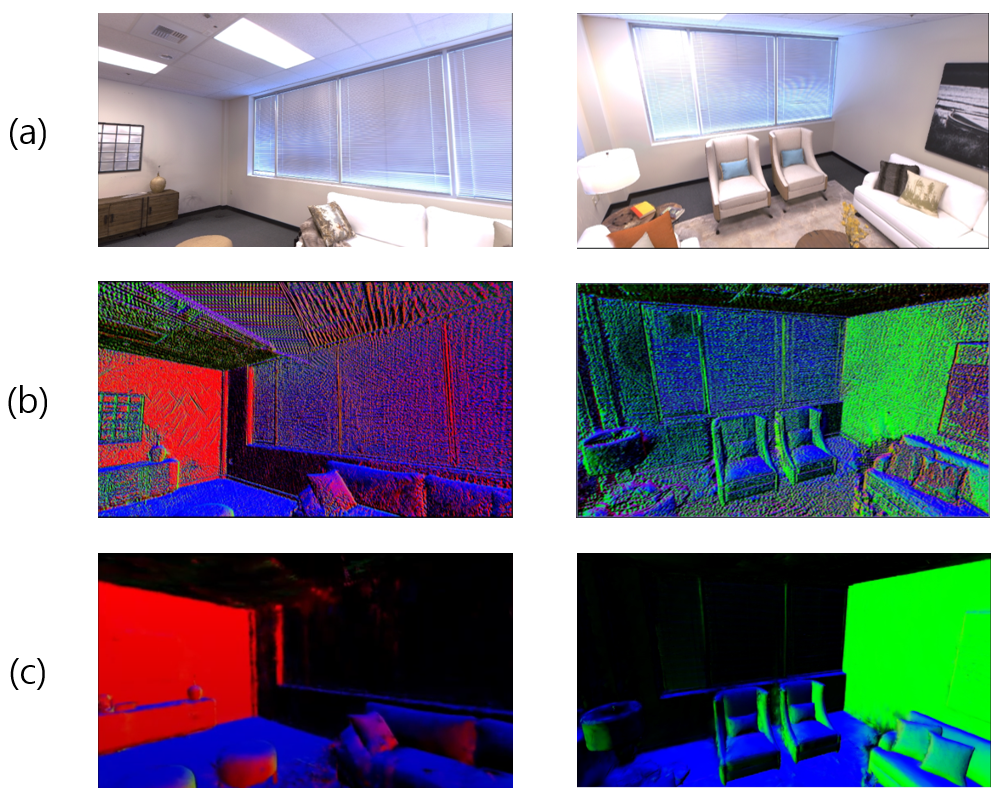}
\caption{Comparison of rendered normal image quality in Replica dataset. Each row is (a) Query image and a rendered normal image of (b) GS-ICP SLAM, (c) G\textsuperscript{2}S-ICP SLAM}
\label{fig_normal_image}
\end{figure}

Building on the strengths of 3DGS, recent studies have explored integrating it into SLAM pipelines. These efforts can be broadly classified into two categories. The first class ~\cite{photoslam, gsicpslam} of approaches focuses on 3DGS-based mapping using externally provided or precomputed poses. While this strategy enables high-fidelity reconstruction with photorealistic rendering, it operates in an open-loop fashion and does not contribute to pose estimation. In contrast, second class ~\cite{splatam, gaussiansplattingslam, gsslam} incorporates 3DGS into tracking, leveraging differentiable rendering or optimization-based methods to estimate camera poses. Despite their potentials, these methods often suffer from high computational cost, sensitivity to lighting conditions, and limited robustness in texture-less or repetitive environments.

These limitations reveal a critical gap: existing 3DGS-SLAM systems prioritize visual quality over geometric stability, making them suboptimal for real-time localization and mapping. To overcome this, we propose a geometry-aware SLAM framework that preserves the visual expressiveness of GS while ensuring spatial consistency and depth fidelity through surface-aligned 2D Gaussian modeling~\cite{2dgs}.

To this end, G\textsuperscript{2}S-ICP SLAM introduces three key innovations. First, we adopt a 2D Gaussian disk representation aligned with the local surface geometry, which solves multi-view depth inconsistencies caused by volumetric 3D Gaussians. Second, we incorporate surface-aligned 2D Gaussian disks into the generalized ICP (GICP)~\cite{segal2009generalized} process by explicitly enforcing anisotropic covariance structures aligned with the local tangent plane. While underlying registration algorithm remains unchanged, our formulation injects 2D priors throuch scale regularization, guiding the optimization toward surface-consistent alignment. Third, we introduce a geometry-aware optimization framework that incorporates photometric, depth, and normal-based geometry-aware losses, along with a scale regularization strategy that preserves surface anisotropy during mapping.

These contributions enable our framework to achieve real-time performance while maintaining geometric consistency and photorealistic rendering quality. Experiments on the TUM-RGBD and Replica datasets demonstrate that G\textsuperscript{2}S-ICP SLAM outperforms prior 3DGS SLAM systems in both localization accuracy and depth fidelity.

\section{Related Work}
\subsection{Radiance Field-based SLAM}
Radiance field-based SLAM methods, including NeRF-SLAM~\cite{orbeez, pointslam, eslam, rosinol2022nerf, wang2023co} and GS-based SLAM~\cite{splatam, gaussiansplattingslam, gsslam,photoslam, gsicpslam}, aim to reconstruct photorealistic scenes by integrating implicit or explicit rendering into SLAM pipelines. NeRF-SLAM leverages implicit volumetric fields to model appearance, while GS-based methods adopt explicit 3D Gaussian splats for efficient rendering and map fusion. Recent works such as SplaTAM~\cite{splatam}, MonoGS~\cite{gaussiansplattingslam}, and GS-SLAM~\cite{gsslam} demonstrate that 3D Gaussian Splatting (3DGS) offers superior visual fidelity and memory efficiency over traditional dense or sparse map representations, making it highly suitable for SLAM applications.

Despite these benefits, 3DGS-based SLAM approaches often suffer from depth inconsistencies due to their volumetric and isotropic nature. Specifically, the unconstrained support along the normal axis in 3D ellipsoids leads to geometry misalignment under multi-view settings. MonoGS\cite{gaussiansplattingslam} addresses some of these issues by incorporating geometric regularization for monocular SLAM, while SplaTAM introduces adaptive map refinement for improved spatial consistency. Meanwhile, Photo-SLAM~\cite{photoslam} and GS-ICP SLAM~\cite{gsicpslam} replace photometric-based pose estimation with classical visual odometry~\cite{orbslam2} or ICP tracking~\cite{segal2009generalized, icp}, enabling faster and more stable operation. However, these methods still inherit limitations of isotropic 3D representations. In contrast, our approach introduces a surface-aligned 2D Gaussian formulation and a geometry-aware loss to address this fundamental bottleneck.

\subsection{Surface-aware Representations for 3D Reconstruction}
Recent advances in radiance field-based mapping, such as 3D Gaussian Splatting (3DGS)~\cite{gaussiansplatting}, have enabled highly detailed and photorealistic scene reconstructions. However, most of these methods use isotropic or volumetric representations—such as 3D Gaussian ellipsoids—that lack explicit surface alignment, often resulting in depth inconsistency and structural artifacts across views. This limitation has led to follow-up works like 2DGS~\cite{2dgs}, SuGaR~\cite{sugar}, and PGSR~\cite{pgsr}, which introduce surface normal supervision or tangent-plane constraints to improve geometric accuracy in neural rendering.

In parallel, geometry-aware learning frameworks have shown that incorporating surface normals as supervision signals significantly improves reconstruction quality. Yet, these insights are not fully integrated into SLAM-oriented systems. Our approach bridges this gap by proposing a surface-aligned 2D Gaussian representation that constrains each Gaussian's support to the local tangent plane, inspired by 2DGS~\cite{2dgs}. Furthermore, we introduce a geometry-aware loss that leverages ground-truth surface normals to guide the rendering process. Together, these components enable our system to preserve fine geometric structure and achieve better reconstruction fidelity without sacrificing runtime or pose tracking performance.

\section{METHODOLOGY}
\begin{figure*}[h]
    \centering
    \includegraphics[width=\linewidth]{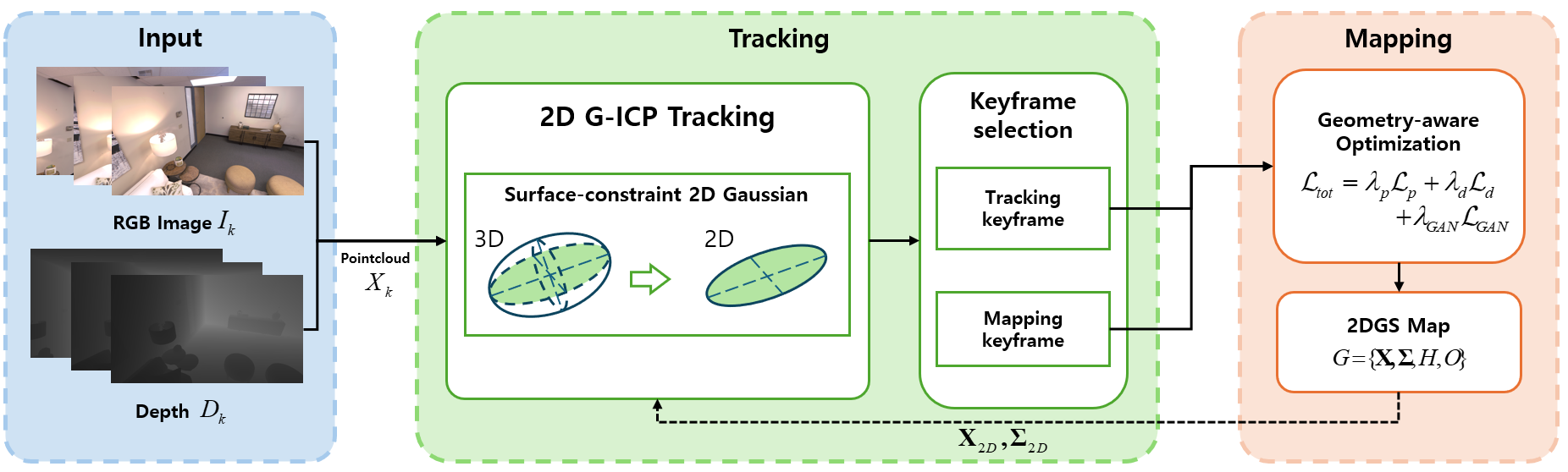}
    \caption{Overview pipeline. 
    G\textsuperscript{2}S-ICP SLAM takes RGB and depth images as input to construct geometrically consistent 3D maps in real time. To enhance multi-view depth consistency, we represent each point as a surface-aligned 2D Gaussian disk instead of a volumetric ellipsoid. In the tracking stage, this representation is integrated into a 2D Gaussian-based Generalized ICP (GICP) to estimate camera motion. During mapping, we optimize the Gaussian primitives using geometry-aware losses that enforce photometric, depth, and normal consistency, thereby preventing orientation misalignments and preserving surface structure.}
    \label{fig:pipeline}
\end{figure*}

This section outlines the core components of G\textsuperscript{2}S-ICP SLAM, a geometry-aware SLAM framework designed to maintain geometric consistency while preserving photorealistic rendering quality. The proposed framework consists of four main components: (1) surface-aligned 2D Gaussian representation, (2) generalized ICP (GICP) tracking with surface-aligned priors, (3) surface aligned scale regularization and (4) geometry-aware optimization. These components bridge the gap between dense photorealistic rendering and spatially consistent SLAM. An overview of the entire system is illustrated in Fig.~\ref{fig:pipeline}, and the detailed implementations follow in the subsequent subsections.

\subsection{Surface-aligned 2D Gaussian Representation}
\label{2dgs}
\begin{figure}[t]
    \centering
    \includegraphics[width=\linewidth]{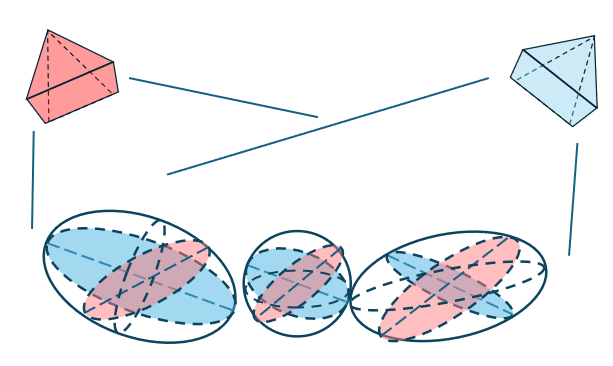}
    \caption{Depth Inconsistency of 3D Gaussian Ellipsoids in Multi-views settings. View-dependent 3D Gaussian ellipsoids cause the same surface point to be represented with different intersection planes across multiple views, resulting in inconsistent depth estimates and geometric misalignment.}
    \label{fig_multiview}
\end{figure}

While exisiting 3DGS SLAM aprroaches adopt 3D Gaussian ellipsoids to represent scenes, their volumetric nature inherently introduces depth inconsistency under multi-view settings~\cite{gsicpslam, photoslam}. Because 3D ellipsoids distribute uncertainty along all three spatial axes, they lack explicit constraints that aligns them with local surface geometry. As a result, the same point may project inconsistently across views, leading to geometric distortion. This issue is illustrated in Fig.~\ref{fig_multiview}, where the same Gaussian ellipsoids are observed from red and blue viewpoints. Each viewpoint intersects the ellipsoid a different slicing plane, resulting in inconsistent depth observations across views and causing depth artifacts in reconstruction.

To address this issue, our method adopts a surface-aligned representation based on 2D Gaussian disks\cite{2dgs}, which constrain the spatial support of each Gaussian to the local tangent plane. This formulation reduces view-dependent ambiguity and improves depth consistency across frames, leading to more stable tracking and geometry-aware SLAM optimization.

Each 2D Gaussian disk is defined by a center position $\boldsymbol{p}_{k}\in\mathbb{R}^{3}$, a covariance matrix $\boldsymbol{C}_{k}\in\mathbb{R}^{3\times3}$, a color $\boldsymbol{c}_{k}$, and an opacity $\sigma_{k}$.
The covariance is computed as:
\begin{equation}
    \boldsymbol{C}_{k} = \boldsymbol{R}_{k}\boldsymbol{S}_{k}\boldsymbol{S}_{k}^{T}\boldsymbol{R}_{k}^{T},
    \label{covariance}
\end{equation}
where $\boldsymbol{R}_{k}=[\boldsymbol{t}_{1}, \boldsymbol{t}_{2}, \boldsymbol{n}_{k}]\in\mathbb{R}^{3\times3}$ is the rotation matrix from two orthogonal tangent vectors and the surface normal, and $\boldsymbol{S}_{k} = \mathrm{diag}(\boldsymbol{s}_{1}, \boldsymbol{s}_{2}, 0)\in \mathbb{R}^{3\times3}$ is the scale matrix aligned to the tangent plane. 

A point $\boldsymbol{p}(u,v)$ on the local tangent plane of the $k$-th Gaussian disk is given by: 
\begin{equation}
\boldsymbol{p}(u, v) = \boldsymbol{p}_{k} + s_{1}t_{1}u + s_{2}t_{2}v = \boldsymbol{H}(u,v,1,1),
\end{equation}
where
\begin{equation}
\boldsymbol{H} = 
\begin{bmatrix}
s_{1}t_{1}&s_{2}t_{2}&0&p_k\\
0 & 0 & 0 & 1\\
\end{bmatrix}
= 
\begin{bmatrix}
\mathbf{RS} & p_k\\
\mathbf{0} & 1\\
\end{bmatrix}.
\end{equation}
The matrix $\boldsymbol{H}_{k}\in\mathbb{R}^{4\times4}$ defines a local surface-aligned transformation that maps 2D disk coordinates onto the 3D global coordinate system. For the point $\boldsymbol{q}(u,v)$ on 2D Gaussian disk is formulated as:
\begin{equation}
    \mathcal{G}(\boldsymbol{q})=\mathrm{exp}{\left({u^{2}+v^{2}}\over{2}\right)}
\end{equation}

We follow a 3DGS rasterization process in 2D Gaussian disk representation. The color and depth rendering is followed as:
\begin{align}
\boldsymbol{c}(\mathbf{x}) = \sum_{i=1}\boldsymbol{c}_{i}\alpha_{i}\mathcal{G}_{i}(\mathbf{x})\prod_{j=1}^{i-1}(1-\alpha_{j}\mathcal{G}_{j}(\mathbf{x})), \\
d(\mathbf{x}) = \sum_{i=1}d_{i}\alpha_{i}\mathcal{G}_{i}(\mathbf{x})\prod_{j=1}^{i-1}(1-\alpha\mathcal{G}_{j}(\mathbf{x})).
\end{align}
This representation inherently captures surface geometry, making the model more robust to depth inconsistencies across multiple views.

\subsection{Generalized ICP Tracking with Surface-aligned Priors}
In the tracking process, we adopt a frame-to-model approach based on Generalized ICP (GICP) registration algorithm. Given an RGB-D image pair $(I_{k}, D_{k})$ at the $k$-th frame, we extract a 3D point cloud $\boldsymbol{\mathcal{X}}_{k} = \{x_{m} \}_{{m=1,...,N_k}}$ from the depth map $D_{k}$, where $N_{k}$ is the number of valid depth pixels. For each point $x_{i}$, we compute a local covariance matrix $\boldsymbol{C}_{m}\in\mathbb{R}^{3\times3}$ by analyzing the spatial distribution of its neighboring points. Using this, we define the Gaussian set for frame $k$ as:
\begin{equation}
    \boldsymbol{G}_{k}= \{\boldsymbol{\mathcal{X}}_{k}, \boldsymbol{\Sigma}_{k} \mid \boldsymbol{\mathcal{X}}_{k} = \{x_{m} \}, \boldsymbol{\Sigma}_{k}=\{\boldsymbol{C}_{m}\}, \forall {m=1,...,N_k}\}.
\end{equation}
Let $\boldsymbol{G}^{src}_{k}= \{\boldsymbol{\mathcal{X}}^{src}_{k}, \boldsymbol{\Sigma}^{src}_{k}\}$ be the Gaussian set extracted from the current frame, and $\boldsymbol{G}^{tgt}_{k}= \{\boldsymbol{\mathcal{X}}^{tgt}_{k}, \boldsymbol{\Sigma}^{tgt}_{k}\}$ be the corresponding target Gaussians set retrieved from the map. The relative transformation $\boldsymbol{T}_{k}\in\mathrm{SE}(3)$ is predicted by minimizing the Mahalanobis distance between corresponding Gaussians.

Each point $\boldsymbol{x}^{src}_{m}$ is modeled as a Gaussian distribution $\mathcal{N}(\hat{\boldsymbol{x}}^{src}_{m}, \boldsymbol{C}^{src}_{m})$, and its corresponding target as $\mathcal{N}(\hat{\boldsymbol{x}}^{tgt}_{m}, \boldsymbol{C}^{tgt}_{m})$. The residual between source and target points is defined as:
\begin{equation}
    d_{m} = \boldsymbol{x}^{tgt}_{m}-\boldsymbol{T}_{k}\boldsymbol{x}^{src}_{m}.
\end{equation}
Assuming the residual follows a Gaussian distribution, we have:
\begin{align}
d_{m} &\sim \mathcal{N}(\hat{d}_{m}, \boldsymbol{C}^{tgt}_{m} + \boldsymbol{T}_{k}\boldsymbol{C}^{src}_{m}(\boldsymbol{T}_{k})^{T}) &\\
&=\mathcal{N}( \hat{x}^{tgt}_{m}- \boldsymbol{T}_{k}\hat{x}^{src}_{m}, \boldsymbol{C}^{tgt}_{m} + \boldsymbol{T}_{k}\boldsymbol{C}^{src}_{m}(\boldsymbol{T}_{k})^{T}).
\end{align}

To obtain the optimal transfromation $\boldsymbol{T}^{*}_{k}$, we apply maximum likelihood estimation across all point correspondences:
\begin{align}
\boldsymbol{T}_{k}^{*} &= \displaystyle \underset{\boldsymbol{T}_{k}}{\mathrm{argmax}}\prod_{m=1}^{N} p(d_{m}) = \underset{\boldsymbol{T}_{k}}{\mathrm{argmax}}\sum_{m=1}^N \mathrm{log} p(d_{m}) \\
&= \displaystyle \underset{\boldsymbol{T}_{k}}{\mathrm{argmax}}\sum_{m=1}^N d_{m}^{T}(\boldsymbol{C}^{tgt}_{m} + \boldsymbol{T}_{k}\boldsymbol{C}^{src}_{m}(\boldsymbol{T}_{k})^{T})^{-1}d_{m}. 
\end{align}

To incorportate surface geometry into this registration process, we embed the anisotropic 2D covariance structure described in Section ~\ref{2dgs} in to each $\boldsymbol{C}^{src}_{m}$. By constraining each covariance matrix to lie on the local tangent plane\---via zero variance along the surface normal direction\---our method introduces an explicit 2D constraint into the GICP tracking. This surface-aligned prior biases the pose estimation toward geometrically consistent alignment, especially under noisy or ambiguous observations.

\subsection{Scale Regularization for 2D Gaussians}
As discussed in Section ~\ref{2dgs}, our framework represents each Gaussian primitive as a 2D disk aligned with the local surface tangent plane. This formulation requires that the spatial uncertainty be constrained to the plane, eliminating variance along the surface normal direction. To satisfy this geometric requirement, we explicitly set the third component of scale matrix $\boldsymbol{S}=diag(s_{1},s_{2},s_{3})$ to zero for all Gaussian primitives used in both tracking and mapping. This configuration ensures that each Gaussian has zero uncertainty along the normal axis, thereby enforcing a flattened 2D representation. 

Unlike prior works that rely on full 3D Gaussian ellipsoids, our system imposes this structural constraint directly during Gaussian construction, avoiding the need for additional regularization terms or optimization overhead. This design guarantees consistent geometric behavior between the tracking module and the mapping representation, both of which share the same set of 2D Gaussians.

\subsection{Distance-aware Gaussian Scaling}
In our mapping module, the global scene is incrementally constructed in a keyframe-centric manner. Two types of keyframes are defined based on distinct criteria: tracking keyframes and mapping keyframes. The tracking keyframes are selected based on the geometric correspondence ratio obtained from the GICP tracking process, while mapping keyframes are inserted at fixed frame intervals to ensure sufficient spatial coverage.

Since the map is built from depth images, the spatial density of the Gaussians is inherently uneven due to the projective characteristics of a range sensor--where each pixel represents a larger spatial area at greater distances. To achieve a photorealistic and perceptually consistent Gaussian representation, we adjust the initial scale of each 2D Gaussian disk according to its distance from the sensor:
\begin{equation}
s_1, s_2 \propto \frac{1}{z^p},
\end{equation}
where $z$ is the depth value and $p$ is a hyperparameter empirically chosen to control the rate of scaling. This distance-aware scaling yields improved visual uniformity and enhances pose estimation stability. 

\begin{table*}[t]
\vspace{-10pt}
\caption{Evaluation of 3D Reconstruction on Replica. Our method outperforms all other real-time Gaussian Splatting SLAM frameworks in quality of the reconstructed map.}
\label{tab:reconstruction_replica}
\centering\resizebox{\linewidth}{!}{
\begin{tabular}{l|c|cccccccccc}
\toprule
Methods & Metrics & R0 & R1 & R2 & Of0 & Of1 & Of2 & Of3 & Of4 & Avg. \\
\midrule
\multirow{4}{*}{\makecell{GS-SLAM~\cite{gsslam}\\(8.34fps)}}
& Depth L1 [cm] $\downarrow$  & 1.31 & 0.82 & 1.26 & 0.81 & 0.96 & 1.41 & 1.53 & 1.08 & 1.16  \\
& Precision [\%] $\uparrow$   & 64.58 & 83.11 & 70.13 & 83.43 & 87.77 & 70.91 & 63.18 & 68.88 & 74.00   \\
& Recall [\%] $\uparrow$      & 61.29 & 76.83 & 63.84 & 76.90 & 76.15 & 61.63 & 62.91 & 61.50 & 67.63   \\
& F1-score [\%] $\uparrow$  & 62.89 & 79.85 & 66.84 & 80.03 & 81.55 & 65.95 & 59.17 & 64.98 & 70.15 \\
\midrule
\multirow{4}{*}{\makecell{GS-ICP SLAM* \\(30fps)}}
& Depth L1 [cm] $\downarrow$  & 22.24 & 17.88 & 23.01 & 16.76 & 16.23 & 23.90 & 21.71 & 22.59 & 20.54  \\
& Precision [\%] $\uparrow$   & 4.77 & 3.69 & 3.63 & 4.77 & 5.20 & 3.54 & 2.50 & 3.58 & 3.96   \\
& Recall [\%] $\uparrow$      & 15.81 & 15.65 & 13.83 & 15.81 & 11.44 & 12.15 & 8.61 & 10.89 & 13.02   \\
& F1-score [\%] $\uparrow$  & 7.33 & 5.97 & 5.75 & 7.33 & 7.15 & 5.49 & 3.87 & 5.39 &  6.04 \\

 \midrule
 \multirow{4}{*}{\makecell{\textbf{G$^{2}$S-ICP SLAM}\\ (30fps)}}
& Depth L1 [cm] $\downarrow$  & \textbf{0.61} & \textbf{0.39} & \textbf{0.90} & \textbf{0.34} & \textbf{0.77} & \textbf{0.98} & \textbf{1.27} & \textbf{0.66} & \textbf{0.74}\\
& Precision [\%] $\uparrow$  & \textbf{85.89} & \textbf{91.47} & \textbf{78.34} & \textbf{90.56} & \textbf{92.35} & \textbf{80.50} & \textbf{76.39} & \textbf{82.23} & \textbf{84.72}    \\
& Recall [\%] $\uparrow$     & \textbf{81.10} & \textbf{84.41} & \textbf{74.24} & \textbf{87.12} & \textbf{83.93} & \textbf{72.29} & \textbf{71.70} & \textbf{74.59} & \textbf{78.67}   \\
& F1-score [\%] $\uparrow$  & \textbf{83.43} & \textbf{87.80} & \textbf{76.23} & \textbf{88.81} & \textbf{87.94} & \textbf{76.17} & \textbf{73.97} & \textbf{78.22} & \textbf{81.57}    \\
\bottomrule
\multicolumn{11}{r}{* denotes the reproduced results by running official code.}\\
\vspace{-10pt}
\end{tabular}}
\end{table*}
\begin{figure*}[t]
    \centering
    \includegraphics[width=\linewidth]{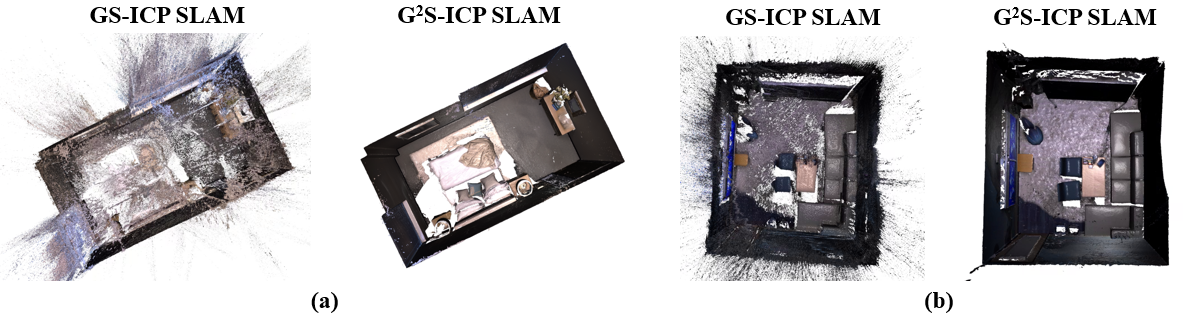}
    \caption{Qualitative Results on Replica dataset. (a), (b) show reconstructed mesh maps of \textit{room1} and \textit{office0} by GS-ICP SLAM and G\textsuperscript{2}S-ICP SLAM. G\textsuperscript{2}S-ICP SLAM achieves clean and accurate 3D mesh reconstruction.}
    \label{fig_reconstrucion}
\end{figure*}

\section{Geometry-aware Optimization}
After pose estimation via 2D GICP tracking and map construction using scale-aware Gaussian initialization, we perform geometry-awrae optimization to refine the Gaussian variables for improved rendering and geometric consistency. Our optimization focuses solely on updating the attributes of the 2D Gaussians to better describe the input observations.

Our total loss function incorporates three complementary term of photometric similarity, depth, geometry-aware normal. 
While photometric and depth losses help preserve visual similarity to ground truth RGB-D observations, the core of our optimization lies in the use of surface normals as a geometric supervisory signal.

We derive surface normals $N_{GT}(x,y)$ from the ground truth depth spatial gradients over 3D points:
\begin{equation}
    N_{GT}(x,y) = {{\triangledown_xx_s\times\triangledown_yx_s}\over{||\triangledown_xx_s\times\triangledown_yx_s||}},
\end{equation}
where $x_s$ denotes the 3D position generated from the depth image. The rendered depth-to-surface normals $\hat{N}_{d}$ are similarly computed from rendered depth $\hat{D}$. The rendered normals $\hat{N}$ are computed from accumulated Gaussian formulation. We then formulate a Geometry-Aware Normal (GAN) loss, which penalizes both the vector difference and angular misalignment among $N$, $\hat{N}$ and $\hat{N}_{d}$:
\begin{align}
\mathcal{L}_{GAN}= \|N_{GT} - \hat{N}\|_1 +\nonumber \left(1 - \frac{N_{GT} \cdot \hat{N}}{\|N_{GT}\|\|\hat{N}\|}\right) \\ + \left(1 - \frac{N_{GT} \cdot \hat{N}_{d}}{\|N_{GT}\|\|\hat{N}_{d}\|}\right).
\end{align}
This normal-based supervision enforces fine-grained surface alignment across viewpoints and complements the auxiliary losses:
\begin{equation}
\mathcal{L}_{p} = \|I - \hat{I}\|_1, \quad \mathcal{L}_{d} = \|D - \hat{D}\|_1,
\end{equation}
which support appearance fidelity and depth regression. The full objective is given by:
\begin{equation}
\mathcal{L} = \lambda_p \mathcal{L}_{p} + \lambda_d \mathcal{L}_{d}. + \lambda_{GAN}\mathcal{L}_{\text{GAN}}
\end{equation}

By integrating surface-aware optimization into the rendering pipeline, our method maintains photorealism while enforcing geometric plausibility—leading to more stable SLAM trajectories and consistent map structures.

\section{Experiments}
\subsection{Expermental Setup}
We describe our experimental setup and compare G\textsuperscript{2}S-ICP SLAM with SOTA mehods to evaluate our performance on 8 synthetic scenes from Replica dataset~\cite{straub2019replica} and 3 scenes from TUM RGB-D dataset~\cite{sturm2012benchmark}.

\noindent\textbf{Implementation Details.}
All experiments were performed on a desktop with Intel Core i5-12500 CPU, 32 GB of RAM, and NVIDIA RTX A5000 GPU. In order to reduce computational resource,  We set the hyper parameters of loss function $\lambda_{P}$, $\lambda_{D}$, and $\lambda_{GAN}$ to 1, 0.1 and 0.05, respectively. We also set the distance-aware scaling factor $p$ to 0.333. To generate mesh of 3D scene, we render and integrate every depth and color frame over the predicted trajectory and use TSDF Fusion~\cite{zeng20163dmatch} with voxel size 1 cm.

\noindent\textbf{Evaluation Metrics.}
For mesh reconstruction, we use the Depth L1~\cite{wang2023co}, which is 2D metric to evaluate scene geometry, Precision, Recall, and F1-score with a distance threshold of 1cm. For tracking accuracy, we evaluate the methods using Root Mean Square Error (RMSE) of the Absolute Trajectory Error (ATE)~\cite{sturm2012benchmark}. We employed standard image quality metrics including Peak Signal-to-Noise Ratio (PSNR), Structural Similarity Index Measure (SSIM)~\cite{wang2004image}, and Learned Perceptual Image Patch Similarity (LPIPS)~\cite{zhang2018perceptual} to evaluate rendering. 

\noindent\textbf{Dataset.}
Replica~\cite{straub2019replica} and TUM-RGBD~\cite{sturm2012benchmark} are the most commonly used datasets for evaluating existing NeRF and GS based SLAM methods. The Replica dataset contains high-quality photorealistic synthetic RGB-D images of various indoor scenes. TUM-RGBD dataset is harder due to lower RGB-D image quality. RGB images contain a motion blur and depth images are noisy and have lots of unfilled regions.

\noindent\textbf{Comparison Methods.}
We consider GS-ICP SLAM~\cite{gsicpslam} as our main baselines for 3D reconstructionm, tracking and rendering performance. We also compare a proposed G\textsuperscript{2}S-ICP SLAM with other SOTA NeRF-based SLAM methods~\cite{nice-slam, pointslam} and other GS-based SLAM~\cite{gsslam, splatam, gaussiansplattingslam}.

\begin{table*}[t]
    \centering
    \caption{Tracking Performance on Replica and TUM-RGBD (ATE RMSE $\downarrow$ [cm]). The best results are highlighted as \colorbox{lightred}{first}, \colorbox{lightorange}{second}, and \colorbox{lightyellow}{third}. Our method achieves state-of-the-art performance in camera pose estimation.}
    \label{tab:ate_replica_tum}
    \resizebox{17.0cm}{!} {
    \begin{tabular}{cl|ccccccccc|cccc} 
    \toprule
    &{\centering Dataset} & \multicolumn{9}{c|}{Replica} & \multicolumn{4}{c}{TUM-RGBD} \\
    &{\centering Method} & R0& R1& R2& Of0& Of1& Of2& Of3& Of4& Avg. & fr1/desk & fr2/xyz & fr3/office & Avg. \\ 
    \midrule
    \multirow{5}{*}{\makecell{Non \\ Real-time}}&NICE-SLAM ~\cite{nice-slam}  & 0.97 & 1.31 & 1.07 & 0.88 & 1.00 & 1.06 & 1.10 & 1.13 & 1.06 &\cellcolor{lightorange}2.7 & 1.8 & 3.0 & 2.5  \\
    &Point-SLAM ~\cite{pointslam} & 0.61 & 0.41 & 0.37 & 0.38 & 0.48 & 0.54 & 0.69 & 0.72 & 0.52 & 4.34 & \cellcolor{lightyellow}1.31 & 3.48 & 3.04 \\
    &GS-SLAM ~\cite{gsslam} & 0.48& 0.53& 0.34& 0.52& 0.41& 0.59& 0.46& 0.70& 0.50 & 3.3 & \cellcolor{lightorange}1.3 & 6.6 & 3.7\\
    &SplaTAM ~\cite{splatam} & \cellcolor{lightyellow}0.31 & 0.40 & \cellcolor{lightyellow}0.29 & 0.47 & \cellcolor{lightyellow}0.27 & 0.29 & 0.32 & \cellcolor{lightyellow}0.55 & \cellcolor{lightyellow}0.36 & 3.35 & \cellcolor{lightred}1.24 & 5.16 & 3.25 \\
    &MonoGS* ~\cite{gaussiansplattingslam} & 0.48& \cellcolor{lightyellow}0.36& 0.34& \cellcolor{lightyellow}0.44& 0.52& \cellcolor{lightyellow}0.23& \cellcolor{lightred}0.16& 2.53& 0.58 & \cellcolor{lightred}1.48 & 1.45 & \cellcolor{lightred}1.50 & \cellcolor{lightred}1.48 \\
    \midrule
    \multirow{2}{*}{\centering Real-time}&GS-ICP SLAM*~\cite{gsicpslam}   &\cellcolor{lightorange} 0.15 & \cellcolor{lightred}0.16 & \cellcolor{lightred}0.10 & \cellcolor{lightorange}0.29 & \cellcolor{lightred}0.12& \cellcolor{lightred}0.16& \cellcolor{lightyellow}0.18& \cellcolor{lightred}0.20& \cellcolor{lightorange}0.17 & 3.07 & 1.79 & \cellcolor{lightorange}2.46 & \cellcolor{lightyellow}2.44\\
    &\textbf{G$^{2}$S-ICP SLAM}   & \cellcolor{lightred}0.14 & \cellcolor{lightred}0.16 & \cellcolor{lightred}0.10 & \cellcolor{lightred}0.19 & \cellcolor{lightred}0.12 & \cellcolor{lightred}0.16 & \cellcolor{lightorange}0.17 & \cellcolor{lightred}0.20 & \cellcolor{lightred}0.15 & \cellcolor{lightyellow}2.74 & 1.59 & \cellcolor{lightyellow}2.78 & \cellcolor{lightorange}2.37\\
    \bottomrule
    \multicolumn{15}{r}{* denotes the reproduced results by running official code.}\\
    \end{tabular}
    }
    
\end{table*}

\subsection{3D Reconstruction Performance}
We evaluate the 3D reconstruction performance of our method on the Replica dataset, as summarized in Table.~\ref{tab:reconstruction_replica}. We compare against GS-SLAM~\cite{gsslam} and GS-ICP SLAM~\cite{gsicpslam} in terms of geometric accuracy and reconstruction completeness. Since our framework achieves real-time operation at 30 FPS, we consider GS-ICP SLAM as a our main baseline, and additionally include GS-SLAM, which provides high-fidelity reconstructions but runs at a significantly lower frame rate of 8.34 FPS.

G\textsuperscript{2}S-ICP SLAM achieves the best geometric accuracy across all scenes, with an average depth L1 error of 0.74 cm, representing 36\% and 96\% improvement over GS-SLAM and GS-ICP SLAM, respectively. This result demonstrates that our surface-aligned 2D Gaussian representation effectively captures 3D structures without introducing geometric noise and enables consistent depth rendering across multiple views.
In addition, our method outperforms all coparisons in reconstruction completeness, reporting the highest average Precision, Recall, and F1-score. These results validate the performance of geometry-aware surface reconstruction for real-time SLAM and confirm that our system produces accurate and complete reconstructions under real-time constraints.

Qualitative results in Fig.~\ref{fig_reconstrucion} further support the superiority of our approach. Compared to GS-ICP SLAM, which often exhibits noisy artifacts and missing structures, G\textsuperscript{2}S-ICP SLAM produces more complete mesh maps. In both room1 and office0 scenes, our method faithfully reconstructs fine-grained geometry and preserves surface continuity, demonstrating robustness in realistic indoor environments.

\subsection{Tracking Performance}

We evaluate the tracking performance of G\textsuperscript{2}S-ICP SLAM on the Replica and TUM-RGBD dataset by comparing it with NeRF-based SLAM~\cite{nice-slam, pointslam} and non real-time GS SLAM~\cite{gsslam, splatam, gaussiansplattingslam} and GS-ICP SLAM~\cite{gsicpslam}. 
As shown in Table.~\ref{tab:ate_replica_tum}, our method consistently achieves the lowest tracking performance on Replica dataset and highly competitive tracking performance in TUM-RGBD dataset. On Replica dataset, G\textsuperscript{2}S-ICP SLAM outperforms all comparison, including GS-ICP SLAM, which is a main baseline, NeRF-based methods, and SOTA GS SLAM systems. On TUM-RGBD, our method achieves the third average results in entire scenes, describing comparative SOTA performance in real-world RGB-D settings.

This strong performance underlines the effectiveness of incorporating 2D surface-aligned constraints into the Generailzed ICP tracking. By replacing conventional 3D ellipsoids with 2D Gaussian disks aligned to local surface geometry, we reduce spatial ambiguity while preserving tracking precision. 
Remarkably, despite imposing a stronger geometric prior, our model shows no degradation in pose estimation accuracy compared to the 3D Gaussian ellipsoid. These results indicate that the 2D constraint not only ensures better geometric consistency but also enhances overall tracking stability in diverse indoor scenes.

\begin{table}
\caption{Evaluation of System Speed and Rendering Quality on Replica. The best results are highlighted as \colorbox{lightred}{first}, \colorbox{lightorange}{second}, and \colorbox{lightyellow}{third}. Our method shows competitive performances both system speed and quality of the rendered image.}
\label{tab:rendering_replica}
\resizebox{8.5cm}{!}
{
\centering
\begin{tabular}{l|c|cccccccccc}
\toprule
Methods & Metrics & R0 & R1 & R2 & Of0 & Of1 & Of2 & Of3 & Of4 & Avg. \\
\midrule
\multirow{4}{*}{GS-SLAM ~\cite{gsslam}}
& PSNR[dB] $\uparrow$    & 31.56 & 32.86 & 32.59 & 38.70     
                         & 41.17 & 32.36 & 32.03 & 32.92 & 34.27         \\
           
& SSIM $\uparrow$        & \cellcolor{lightorange} 0.968 & \cellcolor{lightorange} 0.973 & \cellcolor{lightred} 0.971      &  \cellcolor{lightred}0.986 
           & \cellcolor{lightred} 0.993  & \cellcolor{lightred}0.978 & \cellcolor{lightred}0.970 & \cellcolor{lightorange}0.968 &\cellcolor{lightred} 0.975         \\
           
& LPIPS $\downarrow$     & 0.094  & \cellcolor{lightyellow}0.075 & 0.093 & \cellcolor{lightyellow} 0.050     
                         & \cellcolor{lightred} 0.033 &  0.094 &  0.110 & 0.112 & \cellcolor{lightyellow} 0.082         \\

& FPS $\uparrow$         & 8.34          & -           & -           & -           
           & -          & -           & -           & -           & 8.34
           \\
\midrule

\multirow{4}{*}{SplaTAM ~\cite{splatam}}
& PSNR[dB] $\uparrow$    & 32.86            & 33.89             & 35.25             & 38.26             
                         & 39.17            & 31.97             & 29.70             & 31.81             & 34.11    \\

& SSIM $\uparrow$        & \cellcolor{lightred}0.97 & \cellcolor{lightred}0.98 & \cellcolor{lightorange}0.97 & \cellcolor{lightyellow}0.98              
                         & \cellcolor{lightyellow}0.98 &\cellcolor{lightorange}0.97 & 0.94 & 0.95 & \cellcolor{lightorange}0.97     \\
                         
& LPIPS $\downarrow$     & \cellcolor{lightorange}0.07 & 0.10 & 0.08 & 0.09 
                         & 0.09 & 0.10 & 0.12 & 0.15 & 0.10     \\        

& FPS $\uparrow$         & 0.24 & 0.19 & 0.19 & 0.20 
                         & 0.22 & 0.27 & 0.26 & 0.24 & 0.23     \\
\midrule

\multirow{4}{*}{MonoGS* ~\cite{gaussiansplattingslam}}
& PSNR[dB] $\uparrow$    & \cellcolor{lightorange}33.45 & \cellcolor{lightorange} 36.27 & \cellcolor{lightorange}37.07 & \cellcolor{lightyellow}40.41              
                         & \cellcolor{lightyellow}41.42 & \cellcolor{lightorange} 35.82 & \cellcolor{lightorange} 35.54 & \cellcolor{lightyellow} 33.62 & \cellcolor{lightyellow} 36.70 \\

& SSIM $\uparrow$        & 0.943 & 0.959 & 0.965 & 0.974              
                         & 0.977 & 0.964 & \cellcolor{lightyellow} 0.959 & 0.939 & 0.960     \\
                         
& LPIPS $\downarrow$     & \cellcolor{lightorange} 0.070 & \cellcolor{lightyellow} 0.069 & \cellcolor{lightorange} 0.064 & 0.052             
                         & \cellcolor{lightyellow}0.045 &\cellcolor{lightorange} 0.055 & \cellcolor{lightorange} 0.054 & \cellcolor{lightyellow}0.100  & \cellcolor{lightorange}0.064     \\        

& FPS $\uparrow$         & 0.59 & 0.72 & 0.64 & 0.76 & 0.93 & 0.63 & 0.65 & 0.66 & 0.70     \\
 \midrule
 
\multirow{4}{*}{GS-ICP SLAM*}
& PSNR[dB] $\uparrow$    & \cellcolor{lightred}34.47   & \cellcolor{lightred}36.92    & \cellcolor{lightred}37.37        & \cellcolor{lightred}41.76        
                         & \cellcolor{lightred}42.49   & \cellcolor{lightred}36.01    & \cellcolor{lightred}36.17        & \cellcolor{lightred}38.21        & \cellcolor{lightred}37.92    \\
& SSIM $\uparrow$        & \cellcolor{lightyellow}0.956 & \cellcolor{lightyellow} 0.966 & \cellcolor{lightyellow}0.969 & \cellcolor{lightorange}0.981        
                         & \cellcolor{lightorange} 0.981 & \cellcolor{lightyellow} 0.969 & \cellcolor{lightorange} 0.965        & \cellcolor{lightorange}0.969        & \cellcolor{lightorange}0.970    \\

& LPIPS $\downarrow$     & \cellcolor{lightred}0.057 & \cellcolor{lightred} 0.053  & \cellcolor{lightred} 0.059 & \cellcolor{lightred}0.034        
                         & \cellcolor{lightorange}0.038 & \cellcolor{lightred}0.052 & \cellcolor{lightred} 0.048 & \cellcolor{lightred} 0.051 & \cellcolor{lightred} 0.049    \\
& FPS $\uparrow$         & 29.95            & 29.94        & 29.94        & 29.96        
                         & 29.94            & 29.95        & 29.95        & 29.95        & 29.95    \\
 \midrule

\multirow{4}{*}{\textbf{G$^{2}$S-ICP SLAM}}
& PSNR[dB] $\uparrow$    & \cellcolor{lightyellow}33.40   & \cellcolor{lightyellow}35.73   & \cellcolor{lightyellow}36.12   & \cellcolor{lightorange}40.62   & \cellcolor{lightorange}41.72
                         & \cellcolor{lightyellow}35.32   & \cellcolor{lightyellow}35.14   & \cellcolor{lightorange}37.02   & \cellcolor{lightorange}36.88\\
                         
& SSIM $\uparrow$        & 0.945 & 0.958 & 0.964 & 0.979 & 0.978   
                         & 0.961 & 0.954 & \cellcolor{lightyellow} 0.962 & 0.963 \\

& LPIPS $\downarrow$     & 0.077 & 0.075 & \cellcolor{lightyellow} 0.079 & \cellcolor{lightorange} 0.045 & 0.051 
                         & \cellcolor{lightyellow} 0.069   & \cellcolor{lightyellow} 0.071 & \cellcolor{lightorange} 0.071 & \cellcolor{lightyellow} 0.067   \\
& FPS $\uparrow$         & 29.94   & 29.94   & 29.94   & 29.94    & 29.95
                         & 29.95   & 29.94   & 29.94   & 29.94   \\
\bottomrule
\multicolumn{11}{r}{* denotes the reproduced results by running official code.}\\
\end{tabular}}
\caption{Evaluation of System Speed and Rendering Quality on TUM-RGBD. Proposed method shows incredible system speed and competitive rendering quality.}
\label{tab:rendering_tum}
\centering{
\begin{tabular}{l|cccc}
\toprule
Methods                 
                        & PSNR[dB] $\uparrow$  & SSIM $\uparrow$  & LPIPS $\downarrow$    & FPS $\uparrow$   \\
\midrule

{Photo-SLAM* ~\cite{photoslam}}
                  & \textbf{21.14}           & \textbf{0.738}            & \textbf{0.211}    & -    \\

\midrule

{MonoGS* ~\cite{gaussiansplattingslam}}
                  & 17.91 & 0.716 & 0.311 & 1.77 \\

\midrule
\makecell[l]{GS-ICP SLAM*~\cite{gsicpslam} \\ (limited to 30 FPS)}
                  & 20.42            & 0.764              & 0.226           & \textbf{29.97}     \\

\midrule

\makecell[l]{\textbf{G$^{2}$}\textbf{S-ICP SLAM }\\ (limited to 30 FPS)}
                  & 20.06            & 0.758              & 0.233          & \textbf{29.97}     \\ 
                                            
\bottomrule
\multicolumn{5}{r}{* denotes the reproduced results by running official code.}\\
\end{tabular}}

\end{table}

\subsection{Rendering Performance and System Speed}

\begin{figure*}[t]
    \centering
    \includegraphics[width=\linewidth]{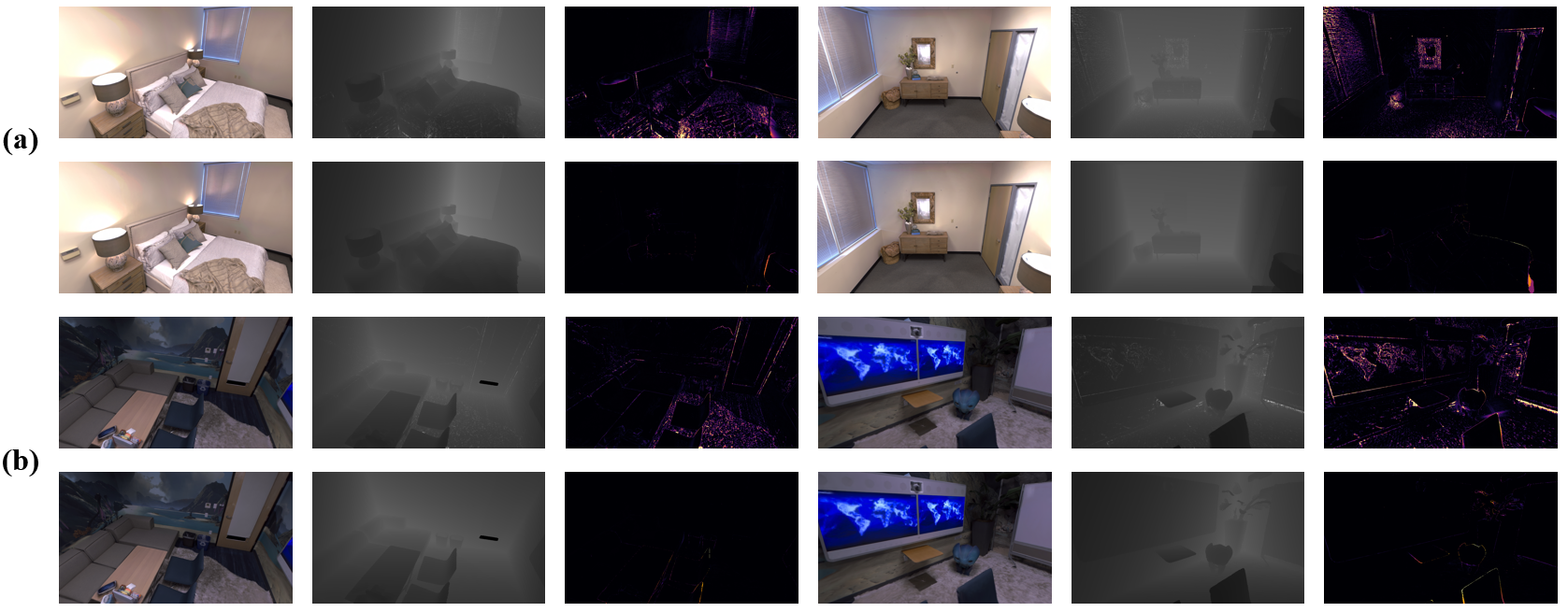}
    \caption{Qualitative Results on Replica dataset. 
    (a), (b) visualize RGB renderings, depth renderings, and depth error maps for two viewpoints in \textit{room1} and \textit{office0} scene. 1st and 3rd rows are rendered by GS-ICP SLAM, 2nd and 4th rows are rendered by G\textsuperscript{2}S-ICP SLAM. G\textsuperscript{2}S-ICP SLAM achieves sharper geometry and lower depth errors.}
    \label{fig_rendering}
\end{figure*}

We evaluate the rendering quality and system speed on the Replica and TUM-RGBD datasets, as summarized in Table~\ref{tab:rendering_replica} and Table~\ref{tab:rendering_tum}. While prior work has empirically shown that 2D Gaussian Splatting~\cite{2dgs} is typically inferior in rendering quality to volumetric 3DGS~\cite{gaussiansplatting}, our results demonstrate that surface-aligned 2D Gaussian modeling can still yield competitive performance. Specifically, G\textsuperscript{2}S-ICP SLAM achieves rendering quality comparable to GS-ICP SLAM, with only a slight drop in rendering performance. This mild degradation is an expected trade-off due to the use of surface-constrained 2D Gaussians, which prioritize geometric alignment over view-dependent appearance blending.
Despite this trade-off, our method outperforms other comparions such as SplaTAM, MonoGS, and GS-SLAM in both rendering quality and efficiency in Replica dataset. Additionally, our method ranks second in PSNR and third in LPIPS on Table~\ref{tab:rendering_replica}.

Fig.~\ref{fig_rendering} further illustrates the advantage of our approach. In both room1 and office0 scenes, G\textsuperscript{2}S-ICP SLAM produces sharper geometry and lower depth error compared to GS-ICP SLAM. The qualitative results show clearer structure boundaries, improved object edges, and fewer missing regions in depth reconstructions. In particular, the depth error maps demonstrate that our method better preserves surface integrity, thanks to the geometry-aware modeling embedded in our SLAM system.

In summary, G\textsuperscript{2}S-ICP SLAM makes a deliberate trade-off between volumetric completeness and geometric precision, offering fast and consistent 3D reconstruction with competitive rendering performance—solidifying the viability of 2D Gaussian splatting in real-time SLAM systems.

\begin{figure}[t]
    \centering
    \includegraphics[width=\linewidth]{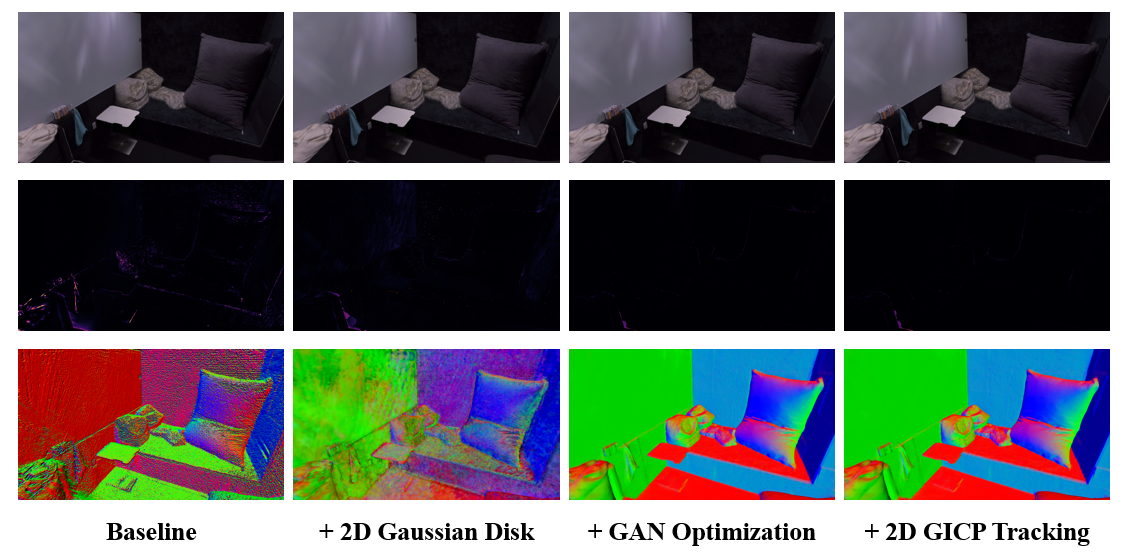}
    \caption{Ablation study on modules. Each row shows the result of rendering image, rendered depth L1 error, rendering normal map.}
    \label{fig:enter-label}
\end{figure}

\begin{table}[t]
    \centering
    \caption{Ablation study on each module in Replica Dataset}
    \begin{tabular}{r|ccccc}
    \toprule
         Method & PSNR[dB] & \makecell{Rendered \\ Depth L1 [cm]}& ATE [cm] \\
         \midrule
         Baseline& 37.92 & 4.180 & 0.17  \\
         \midrule
         + 2D Gaussian Disk & 34.72 & 2.077 & \textbf{0.15} \\
         \midrule
         + GA Optimization & 36.82 & 0.441 & 0.16 \\ 
         \midrule
         Full model& \textbf{36.88} & \textbf{0.437} & \textbf{0.15} \\
         \bottomrule
    \end{tabular}
    \label{tab:ablation_study}
\end{table}

\subsection{Ablation study}

To evaluate the contribution of each component in our framework, we conduct an ablation study on the Replica dataset, as shown in Table~\ref{tab:ablation_study}. Starting from the baseline, we first replace the volumetric 3D ellipsoids with our proposed 2D Gaussian disk representation. This modification alone drastically improves geometric accuracy, reducing the rendered depth L1 error from 4.180 cm to 2.077 cm, and also improves pose estimation accuracy (ATE) from 0.17 cm to 0.15 cm. However, this change comes with a drop in photometric quality (PSNR), highlighting the trade-off between rendering realism and geometric consistency.

Next, we incorporate our geometry-aware optimization loss, which supervises photometric, depth, and surface normal consistency, along with scale regularization. This further reduces the depth error to 0.441 cm while maintaining a competitive ATE of 0.16 cm, and partially recovers the photometric degradation. The full model achieves the best geometric accuracy overall, indicating the complementary benefits of 2D Gaussian alignment and geometry-aware supervision in improving SLAM performance.

These results confirm that each proposed module contributes meaningfully to the final system and that the synergy between surface-aligned representation and geometry-aware loss leads to state-of-the-art performance in both localization and reconstruction fidelity.

\section{Conclusion}
We present G\textsuperscript{2}S-ICP SLAM, a geometry-aware real-time SLAM framework that combines the visual expressiveness of Gaussian Splatting with the spatial consistency of surface-aligned 2D Gaussian representations. By constraining each Gaussian to lie on the local tangent plane, our method solves the depth inconsistencies commonly observed in volumetric 3DGS-based SLAM systems. Furthermore, we introduced a geometry-aware loss formulation that supervises mapping with photometric, depth, and surface normal consistency, while enforcing anisotropic scale regularization to preserve surface alignment during tracking.

Through extensive experiments on the Replica and TUM-RGBD datasets, our method demonstrates significant improvements in both camera localization and 3D reconstruction quality, achieving state-of-the-art performance among GS-based SLAM methods. Notably, our system maintains real-time speed while enhancing depth fidelity and geometric coherence—without compromising rendering quality. These results highlight the effectiveness of incorporating surface-aware priors and geometry supervision into SLAM systems, opening new directions for dense, photorealistic, and structurally accurate scene mapping.


\bibliographystyle{IEEEtran}
\bibliography{bib}

\end{document}